\documentclass[pdflatex,sn-mathphys]{sn-jnl}

\usepackage{subcaption}

\usepackage{hyperref}

\jyear{2023}%

\theoremstyle{thmstyleone}%
\theoremstyle{thmstyletwo}%

\theoremstyle{thmstylethree}%

\begin{document}

\title[ORKG-Leaderboards: A Systematic Workflow for Mining Leaderboards]{ORKG-Leaderboards: A Systematic Workflow for Mining Leaderboards as a Knowledge Graph}


\author*[1]{\fnm{Salomon} \sur{Kabongo}}\email{kabenamualu@l3s.de}

\author[2]{\fnm{Jennifer} \sur{D’Souza}}\email{jennifer.dsouza@tib.eu}

\author[1,2]{\fnm{Sören} \sur{Auer}}\email{soeren.auer@tib.eu}

\affil[1]{\orgdiv{L3S Research Center}, \orgname{Leibniz University of Hannover}, \orgaddress{\city{Hannover}, \state{Lower-saxony}, \country{Germany}}}

\affil[2]{\orgdiv{TIB}, \orgname{Leibniz Information Centre for Science and Technology}, \orgaddress{\city{Hannover}, \state{Lower-saxony}, \country{Germany}}}


\abstract{The purpose of this work is to describe the \textsc{orkg}-Leaderboard software designed to extract \textit{leaderboards} defined as \textit{Task-Dataset-Metric} tuples automatically from large collections of empirical research papers in Artificial Intelligence (AI). The software can support both the main workflows of scholarly publishing, viz. as \LaTeX{} files or as PDF files. Furthermore, the system is integrated with the Open Research Knowledge Graph (ORKG) platform, which fosters the machine-actionable publishing of scholarly findings. Thus the system output, when integrated within the ORKG's supported Semantic Web infrastructure of representing machine-actionable 'resources' on the Web, enables: 1) broadly, the integration of empirical results of researchers across the world, thus enabling transparency in empirical research with the potential to also being complete contingent on the underlying data source(s) of publications; and 2) specifically, enables researchers to track the progress in AI with an overview of the state-of-the-art (SOTA) across the most common AI tasks and their corresponding datasets via dynamic ORKG frontend views leveraging tables and visualization charts over the machine-actionable data. Our best model achieves performances above 90\% F1 on the \textit{leaderboard} extraction task, thus proving \textsc{orkg}-Leaderboards a practically viable tool for real-world usage. Going forward, in a sense, \textsc{orkg}-Leaderboards transforms the \textit{leaderboard} extraction task to an automated digitalization task, which has been, for a long time in the community, a crowdsourced endeavor.}




\maketitle

\section{Introduction}\label{sec1}

Shared tasks---a long-standing practice in the Natural Language Processing (NLP) community---are competitions to which researchers or teams of researchers submit systems that address a specific \textit{Task}, evaluated based on a predefined \textit{Metric}~\cite{sharedtask2}. Seen as ``drivers of progress'' for empirical research, they attract diverse participating groups from both academia and industry, as well as are harnessed as test-beds for new emerging shared tasks on under-researched and under-resourced topics~\cite{sharedtask1}. Examples of long-standing Shared Tasks include the Conference and Labs of the Evaluation Forum (CLEF)\footnote{\url{http://www.clef-initiative.eu/}} organized at the Conference on Natural Language Learning (CoNLL)\footnote{\url{https://www.signll.org/conll}}, the International Workshop on Semantic Evaluation (SEMEVAL)\footnote{\url{https://semeval.github.io/}}, or the biomedical domain-specific BioNLP Shared Task Series~\cite{bionlp} and the Critical Assessment of Information Extraction in Biology (BioCreative)~\footnote{\url{https://biocreative.bioinformatics.udel.edu/tasks/}}. Being inherently competitive, Shared Tasks offer as a main outcome \textit{Leaderboards} that publish participating system rankings. 



Inspired by Shared Tasks, the \textit{Leaderboards} construct of progress trackers is simultaneously taken up for the recording of results in the field of empirical Artificial Intelligence (AI) at large. Here the information is made available via the traditional scholarly publishing flow as PDFs and preprints, unlike in Shared Tasks where the community is relegated to a list of researchers wherein tracking the dataset creators and individual systems applied is less cumbersome as they can be found within the list of researchers that sign up to organize or participate in the task. On the other hand, general publishing avenues bespeak of a deluge of peer-reviewed scholarly publications~\cite{jinha2010article} and PDF preprints ahead (or even instead) of peer-reviewed publications~\cite{chiarelli2019accelerating}. This high-volume publication trend problem is only compounded by the diversity in empirical AI research where \textit{Leaderboards} can potentially be searched and tracked on research problems in various fields such as Computer Vision, Time Series Analysis, Games, Software engineering, Graphs, Medicine, Speech, Audio processing, Adversarial learning, etc. Thus the problem of obtaining completed \textit{Leaderboard} representations of empirical research seems a tedious if not completely insurmountable task. 

Regardless of the setup, i.e. from Shared Tasks or empirical AI research, another problem in the current methodology is the information representation of \textit{Leaderboards} which is often via Github repositories, shared task websites, or researchers' personal websites. Some well-known websites that exist to this end are: PapersWithCode (PwC)~\cite{PWC},\footnote{\url{https://paperswithcode.com}} NLP-Progress~\cite{np}, AI-metrics~\cite{aim}, SQUaD explorer~\cite{squad-exp}, Reddit SOTA~\cite{reddit}. The problem with leveraging websites for storing \textit{Leaderboards} is the resulting rich data's lack of machine-actionability and integrability. In other words, unstructured, non-machine-actionable information from scholarly articles is converted to semi-structured information on the websites which still unfortunately remain non-machine-actionable. In the broader context of scholarly knowledge, the FAIR guiding principles for scientific data management and stewardship~\cite{wilkinson2016fair} identify general guidelines for making data and metadata machine-actionable by making them maximally Findable, Accessible, Interoperable, and Reusable for machines and humans alike. Semantic Web technologies such as the W3C recommendations Resource Description Framework (RDF) and Web Ontology Language (OWL) are the most widely-accepted choice for implementing the FAIR guiding principles~\cite{jacobsen2019fair}. In this context, the Open Research Knowledge Graph (ORKG)~\cite{orkg} \url{https://orkg.org/} as a next-generation library for digitalized scholarly knowledge publishing presents a framework fitted with the necessary Semantic Web technologies to enable the encoding of \textit{Leaderboards} as FAIR, machine-actionable data. Adopting semantic standards to represent \textit{Leaderboards} not just \textit{Task-Dataset-Metric} but also related information such as code links, pre-trained models, and so on can be made machine-actionable and consequently queryable. This would directly address the lack of transparency and integration of various results' problems identified in current methods of recording empirical research~\cite{sharedtask3,sharedtask1,sharedtask2}.

This work, taking note of the two main problems around \textit{Leaderboard} construction, i.e. \textit{information capture} and \textit{information representation}, proposes solutions to address them directly. First, regarding information capture, we recognize due to the overwhelming volume of data, now more than ever, that it is of paramount importance to empower scientists with automated methods to generate the \textit{Leaderboards} oversight. The community could greatly benefit from an automatic system that can  generate a \textit{Leaderboard} as a \textit{Task-Dataset-Metric} tuple over large collections of scholarly publications both covering empirical AI, at large and encapsulating Shared Tasks, specifically. Thus, we empirically tackle the \textit{Leaderboard} knowledge mining machine learning (ML) task via a detailed set of evaluations involving large datasets for the two main publishing workflows, i.e. as \LaTeX{} source and PDF, with several ML models. For this purpose, we extend the experimental settings from our prior work~\cite{kabongo2021automated} by adding support for information extraction from \LaTeX{} code source and compared empirical evaluations on longer input sequences (beyond 512 tokens) for both XLNet and BigBird \cite{zaheer2020big}. Our ultimate goal with this study is to help the Digital Library (DL) stakeholders to select the optimal tool to implement knowledge-based scientific information flows w.r.t. \textit{Leaderboard}s. To this end, we evaluate four state-of-art transformer models, viz. BERT, SciBERT, XLNet, and BigBird, each of which has its own unique strengths. Second, regarding information representation, \textsc{orkg}-Leaderboards workflow, is integrated in the knowledge-graph-based DL infrastructure of the ORKG~\cite{orkg}. Thus the resulting data will be made machine-actionable and served via the dynamic ORKG Frontend views \footnote{\url{https://orkg.org/benchmarks}} and further queryable via structured queries over the larger scholarly KG using SPARQL\footnote{\url{https://orkg.org/triplestore} or \url{https://orkg.org/sparql/}}.



In summary, the contributions of our work are:
\begin{enumerate}
    \item we construct a large empirical corpus containing over 4,000 scholarly articles and 1,548 \textit{leaderboards} TDM triples for the development of text mining systems; 
    \item we empirically evaluate three different transformer models and leverage the best model, i.e. \textsc{orkg}-Leaderboards$_{XLNet}$, for the ORKG benchmarks curation platform;
    \item produced a pipeline that works both with the raw PDF and the \LaTeX{} code source of a research publication. 
    \item we extended our previous work \cite{kabongo2021automated} by empirically investigating our approach with longer input beyond the traditional 512 sequence length limit by BERT-based models, and added support for both mainstreams of research publication PDFs and \LaTeX{} code source.
    \item in a comprehensive empirical evaluation of \textsc{orkg}-Leaderboards for both  \LaTeX{} and PDFs based pipelines, we obtain around 93\% micro and 92\% macro F1 scores which outperform existing systems by over 20 points.
\end{enumerate}

To the best of our knowledge, the \textsc{orkg}-Leaderboards system obtains state-of-the-art results for the \textit{Leaderboard} extraction defined as \textit{Task-Dataset-Metric} triples extraction from empirical AI research articles handling both \LaTeX{} and PDF formats. Thus \textsc{orkg}-Leaderboards can be readily leveraged within KG-based DLs and be used to comprehensively construct \textit{Leaderboards} with more concepts beyond the TDM triples. To facilitate further research, our data\footnote{\url{https://doi.org/10.5281/zenodo.7419877}} and code\footnote{\url{https://github.com/Kabongosalomon/task-dataset-metric-nli-extraction/tree/latex}} are made publicly available.

\section{Definitions}
\label{sec:def}

This section defines the central concepts in the \textit{Task-Dataset-Metric} extraction schema of \textsc{orkg}-Leaderboards. Furthermore, the semantic concepts used in the information representation for the data in the ORKG are defined.

\paragraph{Task.} It is a natural language mention phrase of the theme of the investigation in a scholarly article. Alternatively referred to as research problem~\cite{d2022computer} or focus~\cite{ftd}. An article can address one or more tasks. \textit{Task} mentions being often found in the article Title, Abstract, Introduction, or Results tables and discussion. E.g., question answering, image classification, drug discovery, etc.

\paragraph{Dataset.} A mention phrase of the dataset encapsulates a particular \textit{Task} used in the machine learning experiments reported in the respective empirical scholarly articles. An article can report experiments on one or more datasets. \textit{Dataset} mentions are found in similar places in the article as \textit{Task} mentions. E.g., HIV dataset\footnote{\url{https://wiki.nci.nih.gov/display/NCIDTPdata/AIDS+Antiviral+Screen+Data}}, MNIST~\cite{lecun1998gradient}, Freebase 15K~\cite{bordes2013translating}, etc. 

\paragraph{Metric.} Phrasal mentions of the standard of measurement\footnote{\url{https://www.merriam-webster.com/dictionary/metric}} used to evaluate and track the performance of machine learning models optimizing a \textit{Dataset} objective based on a \textit{Task}. An article can report performance evaluations on one or more metrics. \textit{Metrics} are generally found in Results tables and discussion sections in scholarly articles. E.g., BLEU (bilingual evaluation understudy)~\cite{papineni2002bleu} used to evaluate ``machine translation'' tasks, F-measure~\cite{sasaki2007truth} used widely in ``classification'' tasks, MRR (mean reciprocal rank)~\cite{voorhees1999trec} used to evaluate the correct ordering of a list of possible responses in ``information retrieval'' or ``question answering'' tasks, etc.


\paragraph{Benchmark.} ORKG \textit{Benchmark}s (\url{https://orkg.org/benchmarks}) organize the state-of-the-art empirical research within ORKG \textit{research fields}\footnote{\url{https://orkg.org/fields}} and are powered in part by automated information extraction supported by the \textsc{orkg}-Leaderboards software within a human-in-the-loop curation model. A benchmark per research field is fully described in terms of the following elements: research problem or \textit{Task}, \textit{Dataset}, \textit{Metric}, \textit{Model}, and \textit{Code}. E.g., a specific instance of an ORKG Benchmark \footnote{\url{https://orkg.org/benchmark/R121022/problem/R120872}} on the ``Language Modelling'' \textit{Task}, evaluated on the ``WikiText-2'' \textit{Dataset}, evaluated by ``Validation perplexity'' \textit{Metric} with a listing of various reported Models with respective Model scores.

\paragraph{Leaderboard.} Is a dynamically computed trend-line chart on respective ORKG Benchmark pages leveraging their underlying machine-actionable data from the Knowledge Graph. Thus, \textit{Leaderboard}s depict the performance trend-line of models developed over time based on specific evaluation \textit{Metric}s.

\section{Related Work}

There is a wealth of research in the NLP community on specifying a collection of extraction targets as a unified information-encapsulating unit from scholarly publications. The two main related lines of work that are at the forefront are: 1) extracting instructional scientific content that captures the experimental process~\cite{bioassay,chemrecipes,labprotocols,mysore2019materials,kuniyoshi2020annotating}; and 2) extracting terminology as named entity recognition objectives~\cite{ftd,handschuh2014acl,augenstein2017semeval,Luan2018MultiTaskIO,lrec2020} to generally obtain a concise representation of the scholarly article which also includes the \textit{Leaderboard} information unit~\cite{hou2019identification,scirex,mondal2021end}.

Starting with the capture of the experimental process, \cite{bioassay} proposed an AI-based clustering method for the automatic semantification of bioassays based on the specification of the BAO ontology\footnote{\url{https://github.com/BioAssayOntology/BAO}}.
In~\cite{labprotocols}, they annotate wet lab protocols, covering a large spectrum of experimental biology w.r.t. lab procedures and their attributes including materials, instruments, and devices used to perform specific actions as a prespecified machine-readable format as opposed to the ad-hoc documentation norm. Within scholarly articles, such instructions are typically published in the Materials and Method section in Biology and Chemistry fields. Similarly, in~\cite{chemrecipes,mysore2019materials}, to facilitate machine learning models for automatic extraction of materials syntheses reactions and procedures from text, they present datasets of synthesis procedures annotated with semantic structure by domain experts in Materials Science. The types of information captured include synthesis operations (i.e. predicates), and the materials, conditions, apparatus, and other entities participating in each synthesis step.

In terms of extracting terminology to obtain a concise representation of the article, an early dataset called the FTD corpus~\cite{ftd} defined \textit{focus}, \textit{technique}, and \textit{domain} entity types which were leveraged to examine the influence between research communities. Another dataset, the ACL RD-TEC corpus~\cite{handschuh2014acl} identified seven conceptual classes for terms in the full-text of scholarly publications in Computational Linguistics, viz. \textit{Technology and Method}, \textit{Tool and Library}, \textit{Language Resource}, \textit{Language Resource Product}, \textit{Models}, \textit{Measures and Measurements}, and \textit{Other} to generate terminology lists. Similarly, terminology mining is the task of scientific keyphrase extraction. Extracting keyphrases is an important task in publishing platforms as they help recommend articles to readers, highlight missing citations to authors, identify potential reviewers for submissions, and analyze research trends over time. Scientific keyphrases, in particular, of type \textit{Processes}, \textit{Tasks} and \textit{Materials} were the focus of the SemEval17 corpus annotations~\cite{augenstein2017semeval} which included full-text articles in Computer Science, Material Sciences, and Physics. The SciERC corpus~\cite{Luan2018MultiTaskIO} provided a resource of annotated abstracts in Artificial Intelligence which annotations for six concepts, viz. \textit{Task}, \textit{Method}, \textit{Metric}, \textit{Material}, \textit{Other-Scientific Term}, and \textit{Generic} to facilitate the downstream task of generating a searchable KG of these entities. On the other hand, the STEM-ECR corpus~\cite{lrec2020} notable for its multidisciplinarity included 10 different STEM domains annotated with four generic concept types, viz. \textit{Process}, \textit{Method}, \textit{Material}, and \textit{Data} that mapped across all domains, and further with terms grounded in the real world via Wikipedia/Wiktionary links. Finally, several works have recently emerged targeting the task of Leaderboard extraction, with the TDM-IE pioneering work~\cite{hou2019identification} also addressing the much harder \textit{Score} element as an extraction target. Later works attempted the document-level information extraction task by defining explicit relations \textit{evaluatedOn} between \textit{Task} and \textit{Dataset} elements and \textit{evaluatedBy} between \textit{Task} and \textit{Metric}~\cite{scirex,mondal2021end}. In contrast, in our prior \textsc{orkg}-TDM system~\cite{kabongo2021automated} and in this present extended \textsc{orkg}-Leaderboards experimental report, we attempt the \textit{Task-Dataset-Metric} tuple extraction objective assuming implicitly encoded relations. This simplifies the pipelined entity and relation extraction objectives as a single tuple inference task operating over the entire document. Nevertheless, \cite{scirex,mondal2021end} also defined coreference relations between similar term mentions, which can be leveraged complementarily in our work to enrich the respective \textit{Task-Dataset-Metric} mentions.

\section{The ORKG-Leaderboards Task Dataset}

\subsection{Task Definition}
\label{sec:task}

The \textit{Leaderboard} extraction task addressed in \textsc{orkg}-Leaderboards can be formalized as follows. Let $p$ be a paper in the collection $P$. Each $p$ is annotated with at least one triple $(t_i,d_j,m_k)$ where $t_i$ is the $i^{th}$ \textit{Task} defined, $d_j$ the $j^{th}$ \textit{Dataset} that encapsulates \textit{Task} $t_i$, and $m_k$ is the $k^{th}$ evaluation \textit{Metric} used to evaluate a system performance on a \textit{Task}'s \textit{Dataset}. While each paper has a varying number of \textit{Task-Dataset-Metric} triples, they occur at an average of roughly 4 triples per paper.

In the supervised inference task, the input data instance corresponds to the pair: a paper $p$ represented as the DocTAET context feature $p_{DocTAET}$ and its \textit{Task-Dataset-Metric} triple $(t,d,m)$. The inference data instance, then is $(c; [(t,d,m), p_{DocTAET}])$ where $c \in \{true, false\}$ is the inference label. Thus, specifically, our \textit{Leaderboard} extraction problem is formulated as a natural language inference task between the DocTAET context feature $p_{DocTAET}$ and the $(t,d,m)$ triple annotation. $(t,d,m)$ is $true$ if it is among the paper's \textit{Task-Dataset-Metric} triples, where they are implicitly assumed to be related, otherwise $false$. The $false$ instances are artificially created by a random selection of inapplicable $(t,d,m)$ annotations from other papers. Cumulatively, \textit{Leaderboard} construction is a multi-label, multi-class inference problem.

\subsubsection{DocTAET Context Feature}

The DocTAET context feature representation~\cite{hou2019identification} selects only the parts of a paper where the \textit{Task-Dataset-Metric} mentions are most likely to be found. While the \textit{Leaderboard} extraction task is applicable on the full scholarly paper content, feeding a machine learning model with the full article is disadvantageous since the model will be fed with a large chunk of text which would be mostly noise as it is redundant to the extraction task. Consequently, an inference model fed with large amounts of noise as contextual input cannot generalize well. Instead, the DocTAET feature was designed to heuristically select only those parts of an article that are more likely to contain \textit{Task-Dataset-Metric} mentions as true contextual information signals. Specifically, as informative contextual input to the machine learning model, \underline{DocTAET} captures sentences from four specific places in the article that are most likely to contain \textit{Task-Dataset-Metric} mentions, viz. the \underline{\textbf{Doc}}ument \underline{\textbf{T}}itle, \underline{\textbf{A}}bstract, first few lines of the \underline{\textbf{E}}xperimental setup section and \underline{\textbf{T}}able content and captions.

\subsection{Task Dataset}

To facilitate supervised system development for the extraction of \textit{Leaderboards} from scholarly articles, we built an empirical corpus that encapsulates the task. \textit{Leaderboard} extraction is essentially an inference task over the document. To alleviate the otherwise time-consuming and expensive corpus annotation task involving expert annotators, we leverage distant supervision from the available crowdsourced metadata in the PwC (\url{https://paperswithcode.com/}) KB. In the remainder of this section, we explain our corpus creation and annotation process.

\subsubsection{Scholarly Papers and Metadata from the PwC Knowledge Base.}
We created a new corpus as a collection of scholarly papers with their \textit{Task-Dataset-Metric} triple annotations for evaluating the \textit{Leaderboards} extraction task inspired by the original IBM science result extractor~\cite{hou2019identification} corpus. The collection of scholarly articles for defining our \textit{Leaderboard} extraction objective is obtained from the publicly available crowdsourced leaderboards PwC. It predominantly represents articles in the Natural Language Processing and Computer Vision domains, among other AI domains such as Robotics, Graphs, Reasoning, etc. Thus, the corpus is representative for empirical AI research. The original downloaded collection (timestamp 2021-05-10 at 12:30:21)\footnote{Our corpus was downloaded from the PwC GitHub repository \url{https://github.com/paperswithcode/paperswithcode-data} and was constructed by combining the information in the files \textit{All papers with abstracts} and \textit{Evaluation tables} which included article urls and TDM crowdsourced annotation metadata.} was pre-processed to be ready for analysis. While we use the same method here as the science result extractor, our corpus is different in terms of both labels and size, i.e. number of papers, as many more \textit{Leaderboards} have been crowdsourced and added to PwC since the original work. Furthermore, as an extension to our previous work~\cite{kabongo2021automated} on this theme, based on the two main scholarly publishing workflows i.e. as \LaTeX{} or PDF, correspondingly two variants of our corpus are created and their models respectively developed. 

Recently, publishers are increasingly encouraging paper authors to provide the supporting \LaTeX{} files accompanying the corresponding PDF article. The advantage of having the \LaTeX{} source files is that they contain the original article in plain-text format and thus result in cleaner data in downstream analysis tasks. Our prior \textsc{orkg}-TDM~\cite{kabongo2021automated} model was finetuned only on the parsed plain-text output of PDF articles wherein the plain text was scraped from the PDF which results in partial information loss. Thus, in this work, we modify our previous workflow deciding to tune one model on \LaTeX{} source files as input data, given the increasing impetus of authors also submitting the \LaTeX{} source code; and a second model following our previous work on plain-text scraped from PDF articles.

\begin{enumerate}

\item \textbf{\textit{\LaTeX{} pre-processed corpus.}} To obtain the \LaTeX{} sources, we queried arXiv based on the paper titles from the 5361 articles of our original corpus leveraged to developed \textsc{orkg}-TDM~\cite{kabongo2021automated}. Resultingly, \LaTeX{} sources for roughly 79\% of the papers from the training and test datasets in our original work were obtained. Thus the training set size was reduced from 3,753 papers in the original work to 2,951 papers in this work with corresponding \LaTeX{} sources. Similarly, the test set size was reduced from 1,608 papers in the original work to 1,258 papers in this work for which \LaTeX{} sources could be obtained. Thus the total size of our corpus reduced from 5,361 papers to 4,209 papers. Once the \LaTeX{} sources were respectively gathered for the training and test sets, the data had to undergo one additional step of preprocessing. With the help of pandoc\footnote{\url{https://pandoc.org/}}, latex format files were converted into the XML TEI\footnote{\url{https://tei-c.org/}} markup format files. This is the required input for the heuristics-based script that produces the DocTAET feature. Thus the resulting XML files were then fed as input to the DocTAET feature extraction script. The pipeline to reproduce this process is released in our code repository \footnote{\url{https://github.com/Kabongosalomon/task-dataset-metric-nli-extraction/tree/main/data}}.

\item \textbf{\textit{PDF pre-processed corpus.}} For the 4,209 papers with \LaTeX{} sources, we created an equivalent corpus but this time using the PDF files. This is the second experimental corpus variant of this work. To convert PDF to plain text, following along the lines of our previous work~\cite{kabongo2021automated}, the GROBID parser~\cite{GROBID} was applied. The resulting files in XML TEI markup format were then fed into the DocTAET feature extraction script similar to the \LaTeX{} document processing workflow.

\end{enumerate}

\subsubsection{\textit{Task-Dataset-Metric} Annotations} Since the two corpus variants used in the empirical investigations in this work are a subset of the corpus in our earlier work~\cite{kabongo2021automated}, the 4,209 papers in our present corpus, regardless of the variant, i.e. \LaTeX{} or PDF, retained their originally obtained \textit{Task-Dataset-Metric} labels via distant labeling supervision on the PwC knowledge base (KB).


\subsection{Task Dataset Statistics}

Our overall corpus statistics are shown in \autoref{table:dataset_stats}. The column ``Ours-Prior'' reports the dataset statistics of our prior work~\cite{kabongo2021automated} for comparison purposes. The column ``Ours-Present'' reports the dataset statistics of the subset corpus used in the empirical investigations reported in this paper. The corpus size is the same for both the \LaTeX{} and PDF corpus variants. In all, our corpus contains 4,208 papers split as 2,946 as training data and 1,262 papers as test data. There were 1,724 unique  TDM-triples overall. Note that since the test labels were a subset of the training labels, the unique labels overall can be considered as those in the training data. \autoref{table:dataset_stats} also shows the distinct \textit{Tasks}, \textit{Datasets}, \textit{Metrics} in the last three rows. Our corpus contains 262 \textit{Tasks} defined on 853 \textit{Datasets} and evaluated by 528 \textit{Metrics}. This is significantly larger than the original corpus which had 18 \textit{Tasks} defined on 44 \textit{Datasets} and evaluated by 31 \textit{Metrics}.

\begin{table}[h]
\begin{center}
\begin{minipage}{\textwidth}
\begin{tabular*}{\textwidth}{@{\extracolsep{\fill}}l|cccccc@{\extracolsep{\fill}}}
& \multicolumn{2}{@{}c@{}}{\textbf{Ours-Prior}} & \multicolumn{2}{@{}c@{}}{\textbf{Ours-Present}} & \multicolumn{2}{@{}c@{}}{\textbf{Original}} \\\cmidrule{2-3}\cmidrule{4-5}\cmidrule{6-7}%
 & Train & Test & Train & Test & Train & Test \\
\midrule
Papers               & 3,753  &  1,608 & 2,946 & 1,262 & 170  &  167  \\
``unknown'' annotations &  922 & 380 & 2,359 & 992 & 46 &  45 \\
Total TDM-triples & 11,724 & 5,060 & 9,614 & 4,096 & 327  & 294 \\
Avg. number of TDM-triples per paper & 4.1 & 4.1 & 4.3 & 4.2 &  2.64 & 2.41 \\
Distinct TDM-triples & 1,806 & 1,548 & 1,668 & 1,377 & 78  & 78 \\
Distinct \textit{Tasks}       & 288 & 252 & 262 & 228 & 18  & 18 \\
Distinct \textit{Datasets}    & 908 & 798 & 853 & 714 & 44 & 44 \\
Distinct \textit{Metrics}     & 550 & 469 & 528 & 434 & 31 & 31 \\
\botrule
\end{tabular*}
\caption{Ours-Prior~\cite{kabongo2021automated} vs. Ours-Present vs. the original science result extractor~\cite{hou2019identification} corpora statistics. The ``unknown'' labels were assigned to papers with no TDM-triples after the label filtering stage.}
\label{table:dataset_stats}
\end{minipage}
\end{center}
\end{table}

\subsubsection{DocTAET Context Feature Statistics}

\begin{figure}[H]
\centering
\begin{subfigure}{.48\textwidth}
  \centering
  \includegraphics[width=.9\linewidth]{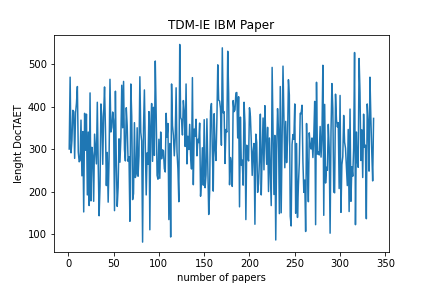}
  \caption{DocTAET feature length in the original science result extractor corpus~\cite{hou2019identification} had a \textbf{max}, \textbf{min}, and \textbf{mean} length of $\textbf{546}$, $\textbf{81}$ and $\textbf{309.45}$, respectively}
  \label{fig:docteat_tdm}
\end{subfigure}\hfill%
\begin{subfigure}{.48\textwidth}
  \centering
  \includegraphics[width=.9\linewidth]{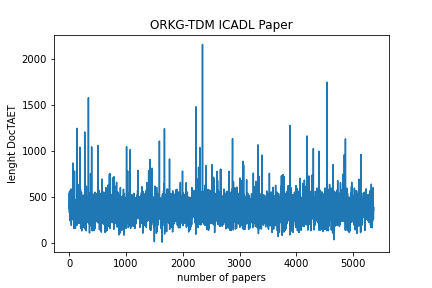}
  \caption{DocTAET feature length in the dataset in our prior work~\cite{kabongo2021automated} had a \textbf{max}, \textbf{min}, and \textbf{mean} length of $\textbf{2161}$, $\textbf{5}$ and $\textbf{378.88}$, respectively}
  \label{fig:docteat_icadl}
\end{subfigure}\hfill
\begin{subfigure}{.48\textwidth}
  \centering
  \includegraphics[width=.9\linewidth]{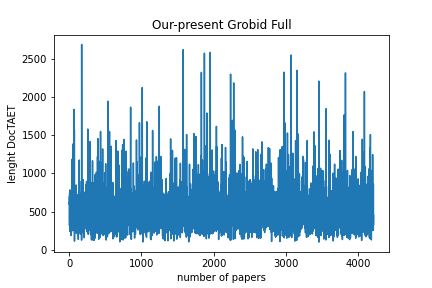}
  \caption{DocTAET feature length in the dataset from the Grobid workflow in our present paper has a \textbf{max}, \textbf{min}, and \textbf{mean} length of $\textbf{2686}$, $\textbf{101}$ and $\textbf{513.37}$, respectively}
  \label{fig:docteat_ours_grobid}
\end{subfigure}\hfill%
\begin{subfigure}{.48\textwidth}
  \centering
  \includegraphics[width=.9\linewidth]{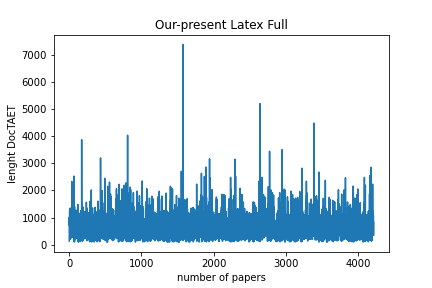}
  \caption{DocTAET feature length in the dataset from the \LaTeX{} workflow in our present paper has a \textbf{max}, \textbf{min}, and \textbf{mean} length of $\textbf{7374}$, $\textbf{100}$ and $\textbf{685.25}$, respectively}
  \label{fig:docteat_ours_latex}
\end{subfigure}
\caption{DocTAET feature length of papers in the original science result extractor dataset~\cite{hou2019identification} \autoref{fig:docteat_tdm}, the dataset used in our prior ORKG-TDM experiements~\cite{kabongo2021automated} \autoref{fig:docteat_tdm}, the dataset from the Grobid workflow in our present work \autoref{fig:docteat_ours_grobid} and the dataset from the \LaTeX{} workflow in our present work \autoref{fig:docteat_ours_latex}.}
\label{fig:docteat}
\end{figure}

\autoref{fig:docteat} shows in detail the variance of the DocTAET Context Feature over three datasets proposed for \textit{Leaderboard} extraction as \textit{Task-Dataset-Metric} triples: 1) \autoref{fig:docteat_tdm} for the dataset from the pioneering science result extractor system~\cite{hou2019identification}; 2) \autoref{fig:docteat_icadl} for the dataset from our prior ORKG-TDM work~\cite{kabongo2021automated}; 3) \autoref{fig:docteat_ours_grobid} and \autoref{fig:docteat_ours_latex} for the dataset in our present paper from the Grobid and \LaTeX{} workflows, respectively (column ``Ours-Present'' in \autoref{table:dataset_stats})).

Both the prior datasets, i.e., the original science result extractor dataset~\cite{hou2019identification} and the ORKG-TDM dataset~\cite{kabongo2021automated}, followed the Grobid processing workflow and reported roughly the same average length of the DocTAET feature. This reflects the consistency preserved in the method of computing the DocTAET feature of between 300 to 400 tokens. Note the ORKG-TDM corpus was significantly larger than the original science result extractor corpus; hence their DocTAET feature length statistics do not match exactly.

In our present paper, as reported earlier, we use a subset of papers from the ORKG-TDM dataset for which the corresponding \LaTeX{} sources could be obtained to ensure similar experimental settings between the Grobid and \LaTeX{} processing workflows. This is why the DocTAET feature length statistics between the ORKG-TDM dataset (\autoref{fig:docteat_icadl}) and our present dataset in the Grobid processing workflow (\autoref{fig:docteat_ours_grobid}) do not match exactly. Still, we see that they are roughly in similar ranges. Finally, of particular interest is observing the DocTAET feature length statistics that could be obtained from the \LaTeX{} processing workflow introduced in this work (\autoref{fig:docteat_ours_latex}). Since from the \LaTeX{} processing workflow cleaner plain-text output could be obtained, the corresponding DocTAET feature lengths in many of the papers were longer than all the rest of the datasets considered, which operated in the Grobid processing workflow over PDFs.

\section{The ORKG-Leaderboards System}

This section depicts the overall end-to-end \textsc{orkg}-Leaderboards, including details on the deep learning models used in our Natural Language Inference (NLI) task formulation. 

\subsection{Workflow}
\begin{figure}[H]
	\centering
    \includegraphics[width=0.6\textwidth]{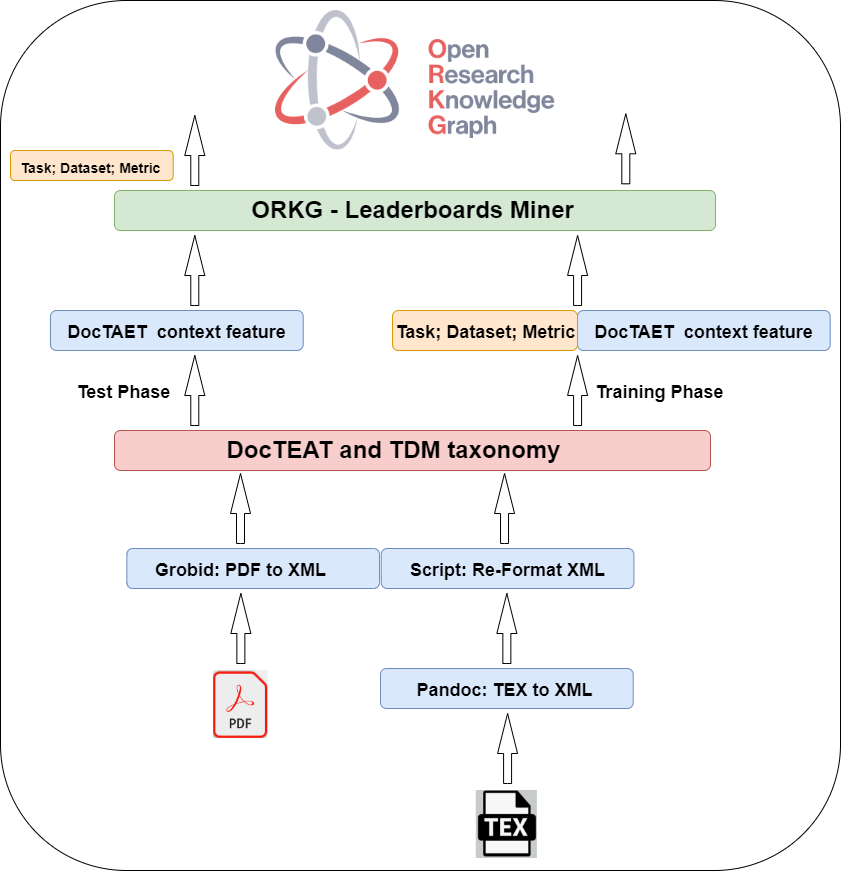}
	\caption{The \textsc{orkg}-Leaderboards end-to-end system workflow in the context of the Open Research Knowledge Graph (ORKG) digital library \url{https://orkg.org/}}
	\label{fig:sys}
\end{figure}

The overall \textsc{orkg}-Leaderboards workflow as depicted in \autoref{fig:sys} includes the following steps:

\begin{enumerate}
    \item A user provides the article input as either the main `.tex' file or a PDF file.
    \item If the input is provided as a `.tex' file, the pandoc script is applied to convert the \LaTeX{} to the corresponding XML TEI marked-up format.
    \item Alternatively, if the input is provided as a PDF file, the Grobid parser is applied to obtain the corresponding scraped plain text in the XML xxxx marked-up format.
    \item Once the XML xxx marked-up files are obtained, the DocTAET feature extraction script is applied to obtain the paper context representations. 
    \item Furthermore, if in the training phase, the collection of papers in the training set is assigned their respective $true$ \textit{Task-Dataset-Metric} labels and a random set of "False" \textit{Task-Dataset-Metric} labels. 
    \item Otherwise, if in the test phase, the query paper is assigned all the \textit{Task-Dataset-Metric} inference targets as candidate labels.
    \item Finally, on the one hand, for the training phase, for each of the input file formats i.e., `.tex' or PDF, an optimal inference model is trained by testing four transformer model variants, viz. BERT, SciBERT, XLNet, and BigBird.
    \item On the hand, for the test phase, depending on the input file format i.e., `.tex' or PDF, the corresponding trained optimal model is applied to the query instance.
    \item Finally, from the test phase, the predicted \textit{Task-Dataset-Metric} tuples output are integrated in the ORKG.
\end{enumerate}

\subsection{Leaderboards Natural Language Inference (NLI)}

To support \textit{Leaderboard} inference~\cite{hou2019identification}, we employ deep transfer learning modeling architectures that rely on a recently popularized neural architecture -- the transformer~\cite{vaswani2017attention}. Transformers are arguably the most important architecture for natural language processing (NLP) today since they have shown and continue to show impressive results in several NLP tasks~\cite{devlin2018bert}. Owing to the self-attention mechanism in these models, they can be fine-tuned on many downstream tasks. These models have thus crucially popularized the transfer learning paradigm in NLP. We investigate three transformer-based model variants for \textit{leaderboard} extraction in a Natural Language Inference configuration. 

Natural language inference (NLI), generally, is the task of determining whether a ``hypothesis'' is true (entailment), false (contradiction), or undetermined (neutral) given a ``premise''~\cite{ref_paperswithcode_nli}. For \textit{leaderboard} extraction, the slightly adapted NLI task is to determine that the (\textit{task}, \textit{dataset}, \textit{metric}) ``hypothesis'' is true (entailed) or false (not entailed) for a paper given the ``premise'' as the DocTAET context feature representation of the paper.

Currently, there exist several transformer-based models. In our experiments, we investigated four core models: three variants of BERT, i.e., the vanilla BERT~\cite{devlin2018bert}, scientific BERT (SciBERT)~\cite{beltagy2019scibert}, and BigBird~\cite{zaheer2020big}. We also tried a different type of transformer model than BERT called XLNet~\cite{yang2019xlnet}, which employs Transformer-XL as the backbone model. Next, we briefly describe the four variants we use.

\paragraph{BERT Models}
BERT (i.e., Bidirectional Encoder Representations from Transformers), is a bidirectional autoencoder (AE) language model. As a pre-trained language representation built on the deep neural technology of transformers, it provides NLP practitioners with high-quality language features from text data simply out of the box and thus improves performance on many NLP tasks. These models return contextualized word embeddings that can be directly employed as features for downstream tasks~\cite{jiang2020improving}.

The first BERT model we employ is BERT\textsubscript{base} (12 layers, 12 attention heads, and 110 million parameters), which was pre-trained on billions of words from the BooksCorpus (800M words) and the English Wikipedia (2,500M words).

The second BERT model we employ is the pre-trained scientific BERT called SciBERT~\cite{beltagy2019scibert}. SciBERT was pretrained on a large corpus of scientific text. In particular, the pre-training corpus is a random sample of 1.14M papers from Semantic Scholar\footnote{\url{https://semanticscholar.org}} consisting of full texts of 18\% of the papers from the computer science domain and 82\% from the broad biomedical field. We used their uncased variants for both BERT\textsubscript{base} and SciBERT.

\paragraph{XLNet}

XLNet is an autoregressive (AR) language model~\cite{yang2019xlnet} that enables learning bidirectional contexts using Permutation Language Modeling. This is unlike BERT's Masked Language Modeling strategy. Thus in PLM, all tokens are predicted but in random order, whereas in MLM, only the masked (15\%) tokens are predicted. This is also in contrast to the traditional language models, where all tokens are predicted in sequential order instead of randomly. Random order prediction helps the model to learn bidirectional relationships and, therefore, better handle dependencies and relations between words. In addition, it uses Transformer XL~\cite{dai2019transformer} as the base architecture, which models long contexts, unlike the BERT models with contexts limited to 512 tokens.
Since only cased models are available for XLNet, we used the cased XLNet\textsubscript{base} (12 layers, 12 attention heads, and 110 million parameters).

\paragraph{BigBird}
BigBird is a sparse-attention-based transformer that extends Transformer based models, such as BERT, to much longer sequences. Moreover, BigBird comes along with a theoretical understanding of the capabilities of a complete transformer that the sparse model can handle~\cite{zaheer2020big}. BigBird takes inspiration from graph sparsification methods by relaxing the need for the attention to fully attend to all the input tokens. Formally the model first builds a set of $g$ global tokens attending on all parts of the sequence, then all tokens attend to a set of $w$ local neighboring tokens, and finally, all tokens attend to a set of $r$ random tokens. The empirical configuration explained in the last paragraph leads to a high-performing attention mechanism scaling to much longer sequence lengths (8x)~\cite{zaheer2020big}. 

\section{ORKG-Leaderboards System Experiments}

\subsection{Experimental Setup}

\paragraph{Parameter Tuning}
We use the Hugging Transformer libraries (\footnote{\url{https://github.com/huggingface/transformers}}) with their BERT variants and XLNet implementations. In addition to the standard fine-tuned setup for NLI, the transformer models were trained with a learning rate of $1e^{-5}$ for 14 epochs; and used the $AdamW$ optimizer with a weight decay of 0 for \textit{bias}, \textit{gamma}, \textit{beta} and 0.01 for the others. Our models' hyperparameters details can be found in our code repository online at \footnote{\url{https://github.com/Kabongosalomon/task-dataset-metric-nli-extraction/blob/main/train_tdm.py}}.

In addition, we introduced a task-specific parameter that was crucial in obtaining optimal task performance from the models. It was the number of $false$ triples per paper. This parameter controls the discriminatory ability of the model. The original science result extractor system~\cite{hou2019identification} considered $|n|-|t|$ \textit{false} instances for each paper, where $|n|$ was the distinct set of triples overall and $|t|$ was the number of $true$ \textit{leaderboard} triples per paper. This approach would not generalize to our larger corpus with over 1,724 distinct triples. In other words, considering that each paper had on average 4 \textit{true} triples, it would have a larger set of \textit{false} triples which would strongly bias the classifier learning toward only \textit{false} inferences. Thus, we tuned this parameter in a range of values in the set \{10, 50, 100\} which at each experiment run was fixed for all papers.

Finally, we imposed an artificial trimming of the DocTAET feature to account for BERT and SciBERT's maximum token length of 512. For this, the token lengths of the experimental setup and table info were initially truncated to roughly 150 tokens, after which the DocTAET feature is trimmed at the right to 512 tokens. Whereas, XLNet and BigBird are specifically designed to handle longer contexts of undefined lengths. Nevertheless, to optimize for training speed, we incorporated a context length of 2000 tokens.

\paragraph{Evaluation}
Similar to our prior work~\cite{kabongo2021automated}, all experiments are performed via two-fold cross-validation. Within the two-fold experimental settings, we report macro- and micro-averaged precision, recall, and F1 scores for our \textit{Leaderboard} extraction task on the test dataset. The macro scores capture the averaged class-level task evaluations, whereas the micro scores represent fine-grained instance-level task evaluations.

Further, the macro and micro evaluation metrics for the overall task have two evaluation settings: 1) considers papers with \textit{Task-Dataset-Metric} and papers with ``unknown'' in the metric computations; and 2) only papers with \textit{Task-Dataset-Metric} are considered while the papers with ``unknown'' are excluded. In general, we focus on the model performances in the first evaluation setting as it directly emulates the real-world application setting that includes papers that do not report empirical research and therefore for which the \textit{Leaderboard} model does not apply. In the second setting, however, the reader still can gain insights into the model performances when given only papers with \textit{Leaderboards}.

\subsection{Experimental Results}
In this section, we discuss new experimental findings shown in \autoref{tab:2 v1}, \autoref{tab:2 v2}, \autoref{tab:best_v1}, and \autoref{tab:best_v2} with respect to four research questions elicited as \textbf{RQ1}, \textbf{RQ2}, \textbf{RQ3}, and \textbf{RQ4} respectively.

\subsubsection*{RQ1: Which is the best model in the real-world setting when considering a dataset of both kinds of papers: those with \textit{Leaderboard}s and those without \textit{Leaderboard}s therefore labeled as ``Unknown''?}
For these results, we refer the reader to the first four results' rows in both \autoref{tab:2 v1} and \autoref{tab:2 v2}, respectively. Note,  \autoref{tab:2 v1} reports results from the Grobid processing workflow and \autoref{tab:2 v2} reports results from the \LaTeX{} processing workflow. In both cases, it can be observed that \textsc{orkg}-Leaderboards$_{XLNet}$ is the best transformer model for the \textit{Leaderboard} inference task in terms of micro-F1. In the case of the Grobid processing workflow, the best micro-F1 from this model is 94.8\%. Whereas in the case of \LaTeX{} processing workflow, the best micro-F1 from \textsc{orkg}-Leaderboards$_{XLNet}$ is 93.0\%. Note in selecting the best model we prefer the micro evaluations since they reflect the fine-grained discriminative ability of the models at the instance level. The macro scores are seen simply as supplementary measures in this regard to observing the performance of the models at the class level.

\subsubsection*{RQ2: How do the models in two processing workflows, i.e. Grobid producing plaintext with some noise and the clean plaintext from \LaTeX{}, compare, both in general and specifically for the best \textsc{orkg}-Leaderboards$_{XLNet}$ model?} 

The model trained on the plain-text obtained from \LaTeX{} contrary to our intuition shows a lower performance compared to the one trained on the noisy Grobid produced plain-text. One possible cause, maybe related to the context length as the \LaTeX{} produced dataset has an average length of $685.25$ compared to $512.37$ for the Grobid produced data, as shown in \autoref{fig:docteat_ours_latex} and \ref{fig:docteat_ours_grobid}. In this case, we hypothesize that for the \LaTeX{} processing workflow to be implemented with the most effective model, experiments with a much larger dataset are warranted. There may be one of two outcomes: 1) the model from the \LaTeX{} workflow still performs worse than the model from the Grobid workflow in which case we can conclude that longer contexts regardless of whether they are from a clean source or noisy source are difficult to generalize from, or 2) the model from the \LaTeX{} workflow indeed begins to outperform the model from the Grobid workflow in which case we can safely conclude that for the transformer models to generalize on longer contexts a much larger training dataset is needed. We relegate these further detailed experiments to future work.




\subsubsection*{RQ3: Which insights can be gleaned from the BERT and SciBERT models operating on shorter context lengths of 512 tokens versus the more advanced models, viz. XLNet and BigBird, operating on longer context lengths of 2000 tokens?}

We observed that BERT and SciBERT models show lower performance compared to the XLNet transformer model operating on 2000 tokens. This we hypothesized as expected behavior since the longer contextual information can capture richer signals for the model to learn from, which is highly likely to be lost when imposing the 512 tokens limit. Contrary to this intuition, however, the BigBird model with the longer context is not able to outperform BERT and SciBERT. We suspect the specific attention mechanism in the BigBird model~\cite{zaheer2020big} needs further examination over a much larger dataset to conclude that it is ineffective for \textit{Task-Dataset-Metric} extraction task compared to other transformer-based models.

\begin{table}[!tb]
\begin{center}
\begin{tabular}{lcccccc}
\hline
\textbf{}        & \textbf{Ma-P}\footnotemark[1] & \textbf{Ma-R} & \textbf{Ma-F1} & \textbf{Mi-P}\footnotemark[2] & \textbf{Mi-R} & \textbf{Mi-F1} \\ \hline
                 & \multicolumn{6}{c}{Average Evaluation Accross 2-fold}                                                                  \\ \hline
\textsc{orkg}-Leaderboards\textsubscript{BERT} &  \textbf{93.2} &	95.7 &	93.5 &	95.4 &	93.9 &	94.7 \\ \hline
\textsc{orkg}-Leaderboards\textsubscript{SciBERT} & 92.6 &	94.3 &	92.2 &	95.4 &	91.1 &	93.2     \\ \hline
\textsc{orkg}-Leaderboards\textsubscript{XLNet} & 93.1 &	\textbf{96.4} &	\textbf{93.7} &	95.1 &	\textbf{94.6} &	\textbf{94.8} \\ \hline
\textsc{orkg}-Leaderboards\textsubscript{BigBird} & \textbf{93.2} &	94.9 &	93.0 &	\textbf{95.7} &	92.4 &	94.0 \\ \hline

\multicolumn{7}{c}{Average Evaluation Across 2-fold (without "Unknown" annotation)} \\
\hline 
\textsc{orkg}-Leaderboards\textsubscript{BERT} & 91.3 &	94.4 &	91.8 &	94.8 &	\textbf{93.9} &	\textbf{94.3}      \\ \hline
\textsc{orkg}-Leaderboards\textsubscript{SciBERT} & 90.5 &	92.5 &	90.3 &	94.8 &	90.6 &	92.7    \\ \hline
\textsc{orkg}-Leaderboards\textsubscript{XLNet} & 91.3 &	\textbf{95.0} &	\textbf{92.0} &	94.3 &	93.5 &	93.9 \\ \hline
\textsc{orkg}-Leaderboards\textsubscript{BigBird} & \textbf{91.5} &	93.3 &	91.3 &	\textbf{95.2} &	92.2 &	93.6	  \\ \hline
\end{tabular}
\renewcommand\thetable{2 v1}
\end{center}
\caption{BERT$_{512}$, SciBERT$_{512}$, XLNet$_{2000}$ and BigBird$_{2000}$ results, trained on the subset of the dataset released by~\cite{kabongo2021automated} from the Grobid workflow}
\footnotetext[1]{Macro Precision}
\footnotetext[2]{Micro Precision.}
\label{tab:2 v1}
\end{table}

\begin{table}[!tb]
\begin{center} \scriptsize
\begin{tabular}{lcccccc}
\hline
\textbf{}        & \textbf{Ma-P}\footnotemark[1] & \textbf{Ma-R} & \textbf{Ma-F1} & \textbf{Mi-P}\footnotemark[2] & \textbf{Mi-R} & \textbf{Mi-F1} \\ \hline
                 & \multicolumn{6}{c}{Average Evaluation Across 2-fold}                                                                  \\ \hline
\textsc{orkg}-Leaderboards\textsubscript{BERT} &  \textbf{93.5} &	94.2 &	\textbf{92.8} &	\textbf{96.0} &	90.0 &	92.9 \\ \hline
\textsc{orkg}-Leaderboards\textsubscript{SciBERT} & 91.7 &	93.9 &	91.6 &	94.6 &	88.6 &	91.5     \\ \hline
\textsc{orkg}-Leaderboards\textsubscript{XLNet} &  91.9 &	\textbf{94.4} &	92.0 &	94.9 &	\textbf{91.2} &	\textbf{93.0} \\ \hline
\textsc{orkg}-Leaderboards\textsubscript{BigBird} & 90.7 &	91.6 &	89.7 &	94.6 &	87.2 &	90.7 \\ \hline

\multicolumn{7}{c}{Average Evaluation Accross 2-fold (without "Unknown" annotation)} \\
\hline 
\textsc{orkg}-Leaderboards\textsubscript{BERT} & \textbf{91.2} &	92.3 &	\textbf{90.6} &	\textbf{95.4} &	88.0 &	91.5      \\ \hline
\textsc{orkg}-Leaderboards\textsubscript{SciBERT} & 89.4 &	91.7 &	89.2 &	93.7 &	86.0 &	89.7    \\ \hline
\textsc{orkg}-Leaderboards\textsubscript{XLNet} &  89.5 &	\textbf{92.4} &	89.8 &	94.2 &	\textbf{89.4} &	\textbf{91.7}  \\ \hline
\textsc{orkg}-Leaderboards\textsubscript{BigBird} & 87.5 &	88.7 &	86.6 &	93.6 &	85.3 &	89.3   \\ \hline
\end{tabular}
\end{center}
\renewcommand\thetable{2 v2}
\caption{BERT$_{512}$, SciBERT$_{512}$, XLNet$_{2000}$ and BigBird$_{2000}$ results, based on DocTEAT from  \LaTeX{} code source.}
\label{tab:2 v2}
\footnotetext[1]{Macro Precision}
\footnotetext[2]{Micro Precision.}
\end{table}

\begin{table}[!tb]
\centering \scriptsize
\begin{tabular}{lcccccccccccc}
\hline\multirow{2}{*} {\textbf{Entity} } & \multicolumn{5}{c} { \textbf{Macro} } & & \multicolumn{5}{c} { \textbf{Micro} } \\
\cline { 2 - 6 } \cline { 8 - 12 } & $\mathrm{P}$ && $\mathrm{R}$ && $\mathrm{F}_{1}$ & & $\mathrm{P}$ && $\mathrm{R}$ && $\mathrm{F}_{1}$ \\
\hline TDM & 93.1 && 96.4 && 93.7 && 95.1 && 94.6 && 94.8  \\
\hline
Task & 94.3 &&	97.2 &&	95.0 &&	96.8 &&	95.9 &&	96.4  \\
Dataset &  93.8 &&	96.7 &&	94.4 &&	96.2 &&	95.4 &&	95.8 \\
Metric & 93.7 &&	96.9 &&	94.4 &&	96.0 &&	95.3 &&	95.6 \\
\hline
\end{tabular}
\renewcommand\thetable{3 v1}
\caption{Performance of our best model, i.e. \textsc{orkg}-Leaderboards\textsubscript{XLNet}, for \textit{Task}, \textit{Dataset}, and \textit{Metric} concept extraction of the \textit{leaderboard} for the grobid workflow}
\label{tab:best_v1}
\end{table}

\begin{table}[!tb]
\centering \scriptsize
\begin{tabular}{lcccccccccccc}
\hline\multirow{2}{*} {\textbf{Entity} } & \multicolumn{5}{c} { \textbf{Macro} } & & \multicolumn{5}{c} { \textbf{Micro} } \\
\cline { 2 - 6 } \cline { 8 - 12 } & $\mathrm{P}$ && $\mathrm{R}$ && $\mathrm{F}_{1}$ & & $\mathrm{P}$ && $\mathrm{R}$ && $\mathrm{F}_{1}$ \\
\hline TDM &  91.9	&&	94.4	&&	92.0	&&	94.9	&&	91.2	&&	93.0  \\
\hline
Task & 94.3	&&	97.2	&&	95.0	&&	96.8	&&	95.9	&&	96.4 \\
Dataset &  93.8	&&	96.7	&&	94.4	&&	96.2	&&	95.4	&&	95.8  \\
Metric &  93.7	&&	96.9	&&	94.4	&&	96.0	&&	95.3	&&	95.6  \\
\hline
\end{tabular}
\renewcommand\thetable{3 v2}
\caption{Performance of our best model, i.e. \textsc{orkg}-Leaderboards\textsubscript{XLNet}, for \textit{Task}, \textit{Dataset}, and \textit{Metric} concept extraction of the \textit{leaderboard} for the latex workflow}
\label{tab:best_v2}
\end{table}

\subsubsection*{RQ4: Which of the three \textit{Leaderboard} \textit{Task-Dataset-Metric} concepts are easy or challenging to extract?}

As a fine-grained examination of our best model, i.e. \textsc{orkg}-Leaderboards\textsubscript{XLNet}, we examined its performance for extracting each of three concepts $(Task, Dataset, Metric)$ separately. These results are shown in \autoref{tab:best_v1} and \autoref{tab:2 v2}. From the results, we observe that \textit{Task} is the easiest concept to extract, followed by \textit{Metric}, and then \textit{Dataset}. We ascribe the low performance for extracting the \textit{Dataset} concept due to the variability in its naming seen across papers even when referring to the same real-world entity. For example, the real-world dataset entity `CIFAR-10' is labeled as `CIFAR-10, 4000 Labels' in some papers and `CIFAR-10, 250 Labels' in others. This phenomenon is less prevalent for \textit{Task} and the \textit{Metric} concepts. For example, the \textit{Task} `Question Answering' is rarely referenced differently across papers addressing the task. Similarly, for \textit{Metric}, `accuracy' as an example, has very few variations.

\begin{figure}[H]
	\centering
	\includegraphics[width=1.0\linewidth]{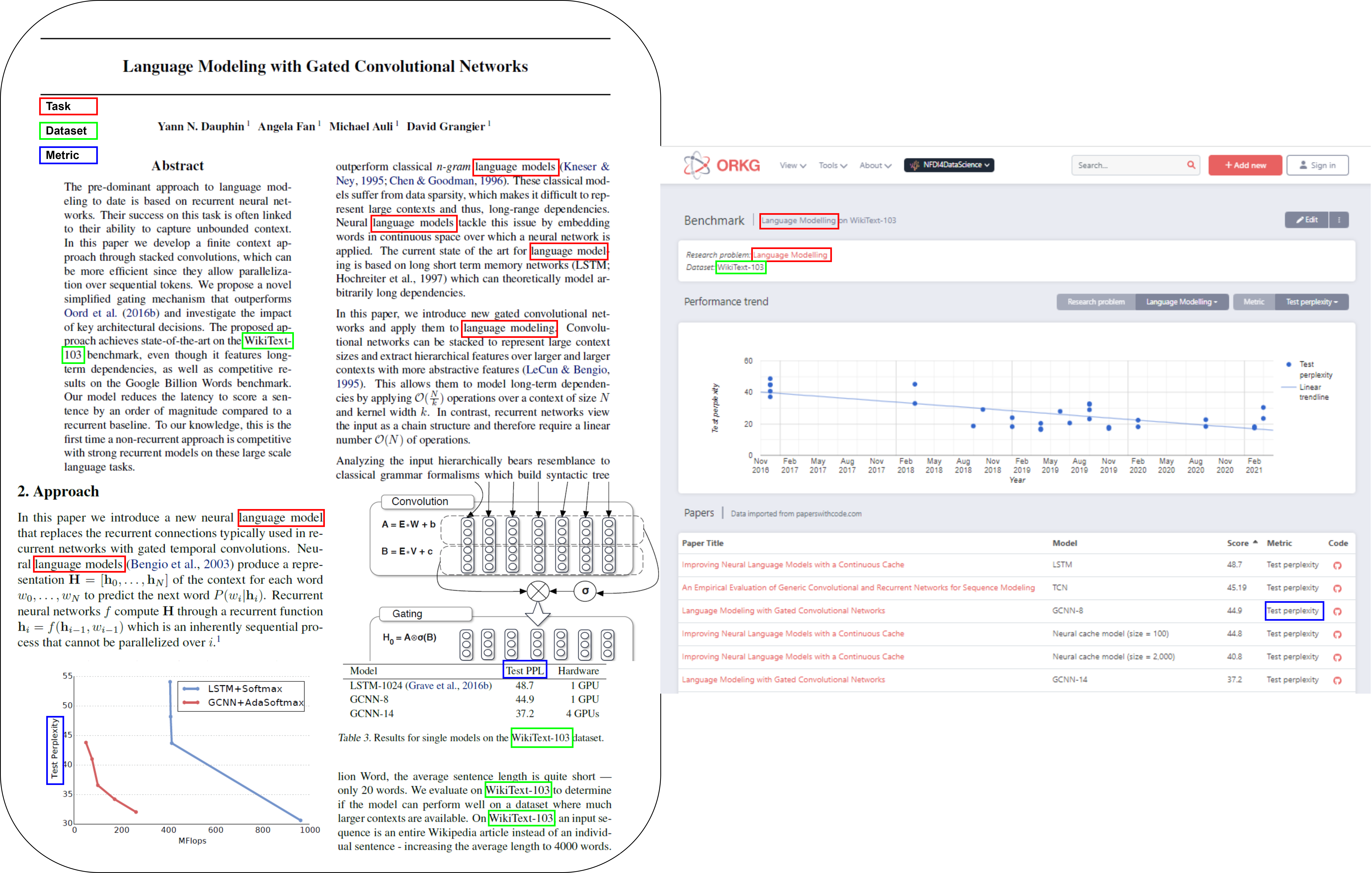}
    \caption{A contrastive view of \textit{Task-Dataset-Metric} information in the traditional PDF format of publishing as non-machine-actionable data (on the left) versus as machine-actionable data with \textit{Task-Dataset-Metric} annotations obtained from \textsc{orkg}-Leaderboards and integrated in the next-generation scholarly knowledge platform as the ORKG Benchmarks view (on the right).}
    \label{fig:dl}
\end{figure}

\section{Integrating ORKG-Leaderboards in the Open Research Knowledge Graph}

In this era of the publications deluge worldwide~\cite{jinha2010article,chiarelli2019accelerating,stm}, researchers are faced with a critical dilemma: \textit{How to stay on track with the past and the current rapid-evolving research progress?} With this work, our main aim is to propose a solution to this problem. And with the \textsc{orkg}-Leaderboards software, we have concretely made advances toward our aim in the domain of empirical AI research. Furthermore, with the software integrated into the next-generation digitalized publishing platform, viz. \url{https://orkg.org/}, the machine-actionable \textit{Task-Dataset-Metric} data represented as a Knowledge Graph with the help of the Semantic Web's RDF language makes the information skimmable for the scientific community. This is achieved via the dynamic Frontend views of the ORKG Benchmarks feature \url{https://orkg.org/benchmarks}. This is illustrated via \autoref{fig:dl}. On the left side of \autoref{fig:dl} is shown the traditional PDF-based paper format. Highlighted within the view are the \textcolor{red}{Task}, \textcolor{green}{Dataset}, and \textcolor{violet}{Metric} phrases. As evident, the phrases are mentioned in several places in the paper. Thus in this traditional model of publishing via non-machine-actionable PDFs, a researcher interested in this critical information would need to scan the full paper content. They are then faced with the intense cognitive burden of repeating such a task over a large collection of articles. On the right side of the \autoref{fig:dl} is presented a dynamic ORKG Frontend view of the same information, however over machine-actionable RDF semantically represented information of the \textcolor{red}{Task}, \textcolor{green}{Dataset}, and \textcolor{violet}{Metric} elements. To generate such a view, the \textsc{orkg}-Leaderboard software would simply be applied on a large collection of articles either in \LaTeX{} or PDF format, and the resulting \textit{Task-Dataset-Metric} tuples uploaded in the ORKG. Note, however, \textsc{orkg}-Leaderboard does not attempt extraction of the \textit{Score} element. We observed from some preliminary experiments that the \textit{Score} element poses a particularly hard extraction target. This is owing to the fact that the underlying contextual data supporting \textit{Score} extraction is especially noisy--clean table data extraction from PDFs are a challenging problem in the research community that would need to be addressed first to develop promising \textit{Score} extractors. Nevertheless, in the context of this missing data in the ORKG Benchmarks pages, its human-in-the-loop curation model is relied on. In such a setting, respective article authors with their \textit{Task-Dataset-Metric} model information being automatically extracted to the KG can simply edit their corresponding model scores in the graph. Thus as concretely shown on the right screen of \autoref{fig:dl}, empirical results are made skimmable and easy to browse for researchers interested in gaining an overview of empirical research progress via a ranked list of papers proposing models and a performance progress trend chart computed over time.



Although the experiments of our study targeted empirical AI research, we are confident, that the approach is transferable to similar scholarly knowledge extraction tasks in other domains. For example in Chemistry or Material Sciences, experimentally observed properties of substances or materials under certain conditions could be obtained from various papers.


\section{Conclusion and Future Work}
In this work we experimented with the empirical construction of Leaderboards, using four recent transformer-based models (BERT, SciBERT, XLNet, BigBird) that have achieved state-of-the-art performance in several tasks and domains in the literature. Leveraging the two main streams of information acquisition used in scholarly communication i.e (Pdf, \LaTeX{}), our work published two models to accurately extract Task, Dataset, and Metric entities from an empirical AI research publication. Therefore as a next step, we will extend the current triples (task, dataset, metric) model with additional concepts that are suitable candidates for a Leaderboard such as score or code URLs, etc. We also envision the task-dataset-metric extraction approach to be transferable to other domains (such as materials science, engineering simulations, etc.). Our ultimate target is to create a comprehensive structured knowledge graph tracking scientific progress in various scientific domains, which can be leveraged for novel machine-assistance measures in scholarly communication, such as question answering, faceted exploration, and contribution correlation tracing.

\subsubsection*{Acknowledgments}
This work was co-funded by the Federal Ministry of Education and Research (BMBF) of Germany for the project LeibnizKILabor (grant no. 01DD20003), BMBF project SCINEXT (GA ID: 01lS22070), NFDI4DataScience (grant no. 460234259) and by the European Research Council for the project ScienceGRAPH (Grant agreement ID: 819536).
\bibliography{main}


\end{document}